\documentclass[]{eptcs}
\providecommand{\event}{ICLP DC 2019} 
\usepackage{breakurl}             
\usepackage{amsmath}
\usepackage{amssymb}
\usepackage{amsfonts}
\usepackage{mathtools}
\usepackage{mathptmx}
\usepackage{listings}
\usepackage{tikz}
\usepackage{graphicx}
\usepackage{subfig}
\usepackage{tikzscale}
\usepackage{float}
\usepackage{backnaur}
\usepackage{alltt}
\newfloat{listing}{thp}{lob}
\floatname{listing}{Listing}

\title{Research Report on Automatic Synthesis of\\ Local Search Neighborhood 
Operators}
\author{Mateusz \'Sla\.zy\'nski
\institute{AGH University of Science and Technology, Poland}
\email{mslaz@agh.edu.pl}
}
\def\titlerunning{Automatic Synthesis of Local Search Neighborhoods}
\def\authorrunning{M. \'Sla\.zy\'nski}

\begin{document}
\maketitle

\begin{abstract}
Constraint Programming (CP) and Local Search (LS) are different
paradigms for dealing with combinatorial search and optimization
problems. Their complementary features motivated researchers to
create hybrid CP/LS solutions, maintaining both the modeling
capabilities of CP and the computational advantages of the
heuristic-based LS approach.  Research presented in this report is focused on 
developing a novel method to infer an efficient LS neighborhood operator based on 
the problem structure, as modeled in the CP paradigm. 
We consider a limited formal language that we call a Neighborhood Definition
Language, used to specify the neighborhood operators in a
fine-grained and declarative manner. Together with Logic Programming runtime 
called Noodle, it allows to automatically synthesize complex operators using a 
Grammar Evolution algorithm.

\end{abstract}

\section{Introduction and problem description}

The Discrete Optimization domain includes a wide class of problems
ranging from planning and scheduling to packing and automated design.
Due to the vast applicability, a number of techniques and tools have been
proposed, to model and solve those problems in a general and
standardized way.  Constraint Programming (CP) and Local Search (LS)
are different paradigms for dealing with combinatorial search and
optimization problems; each of them having different strong and weak
points.

Constraint Programming, rooted in declarative programming, offers
excellent tools to create intelligible and fine-grained models.
Thanks to a small semantic gap between the problem at hand and its
model, it is possible to exploit a problems' structure and work
directly with the domain experts.

On the other hand, most CP solvers rely on a backtracking search
process to explore the solution tree, which does not perform well on
hard optimization problems.  The other disadvantage of the popular
search strategies is a lack of proper relaxation methods (compared to,
e.g. Mixed Integer Programming), which denies search-space pruning
techniques, for instance via the branch and bound approach.

At the same time, LS focuses on finding a \emph{good enough}
solution in a realistic time frames even for difficult real-life
problems.  Moreover, LS can easily be performed in parallel,
which makes it an attractive proposal for modern hardware.

Unfortunately, efficiency of the LS-based solvers depends highly on
the problem's representation and as for today, LS problems are still
mostly modeled directly as computer programs in general-purpose
languages.  Due to this lack of model layer, creating a LS
solver can be a time-consuming effort requiring programming experience
and lot of experimental work.  Moreover, the LS models
inherit all issues of general computer programs, i.e.~they do not
compose well and are difficult to maintain as the problem statement
evolves, due to the large semantic gap.

Those complementary features have motivated researchers to create
hybrid CP/LS solutions, maintaining both the modeling capabilities of
CP and the computational advantages of the heuristic-based LS
methods. The presented research is yet another approach to achieve this goal.

\section{Background and overview of the existing literature}

Frequently, LS solvers are implemented in an imperative
programming language without any modeling layer.  To facilitate code
reuse, several programming libraries were introduced, such as
JAMES~\cite{debeukelaer_james:_2017},
Local++~\cite{schaerf_local++:_1999} or
EasyLocal++~\cite{gaspero_easylocal++:_2003}.  Neighborhood operators
are represented as a class or function satisfying a predefined
interface.  OptaPlanner~\cite{noauthor_optaplanner_nodate} is a hybrid
solver configurable by means of XML files.  It supports several search
strategies, including LS, and allows the definition of the
neighborhood in a declarative manner as a composition of basic moves.
There are several predefined moves applicable to common optimization
problems, but to add new ones, one has to implement them in the Java
language.

At the same time, the Constraint Programming and Mathematical
Optimization communities have been concerned with problem modeling,
which results in several human readable representations.  A
Mathematical Programming Language (AMPL)~\cite{fourer_modeling_1990}
is a modeling language closely resembling algebraic notation that can
be easily translated to internal representation of various
mathematical solvers.  The Optimization Programming Language
(OPL)~\cite{van_hentenryck_opl_1999} provides a more generic
representation supporting both mathematical and constraint programming
tools.  Recently, various Constraint Programming languages have been
proposed, such as Essence~\cite{frisch_essence:_2008} that
incorporates various data structures known from programming languages
or MiniZinc~\cite{bessiere_minizinc:_2007} that is, by design, solver
agnostic thanks to an intermediate low level representation called
FlatZinc.  These tools enable the user to express a problems'
structure in a way as close as possible to their natural form.

Localizer~\cite{michel_localizer_1997} is a language designed to model
the whole LS routine in one domain-specific modeling
language.  The Comet~\cite{van_hentenryck_constraint-based_2005}
language extended the Localizer approach by adding mechanisms to
independently express both a problems' structure and the search
strategy in terms of constraint programming models.  It was possible
to define simple neighborhoods in an imperative manner, by selecting
variables and their new values according to various heuristics.  While
the Comet language is not supported anymore, systems such as
OscaR~\cite{oscar_team_oscar:_2012} provide very similar capabilities.
Neighborhood combinators~\cite{landtsheer_combining_2018} extend OscaR
with a declarative language enabling the user to create search
strategies by combining existing neighborhood operators.  The
neighborhood operators themselves still have to be predefined and
implemented in the Scala language.  A similar, although less
extendable, proposal has been also made for the MiniZinc modeling
language~\cite{pesant_minisearch:_2015}.

Declarative Neighborhoods~\cite{bjordal_declarative_2018} are designed
to improve this situation by providing a MiniZinc extension defining
the neighborhood operators in a declarative manner.  The neighborhood
operator is represented as a constraint satisfaction problem itself,
with constraints corresponding to relations between the neighbors.
Such a representation integrates well into a Constraint Programming
model, allowing one to put additional requirements on the neighborhood
relation, i.e.~every neighbor has to satisfy a set of constraints or
that the neighborhood operator requires some constraints to be
satisfied before it may be applied.

Recently, there has been a proposal for a system inferring possible
neighborhood operators from data structures occurring in the
Constraint Programming model, defined in the Essence 
language~\cite{akgun_framework_2018}.  Various data structures are connected
with relevant move operators that are tried in an intelligent manner
using classic multi-arm bandit strategies.  The resulting solver
proved to be superior to or competitive with popular generic
strategies on several common combinatorial problems.  As the authors
focused more on exploiting basic moves in an unsupervised manner, the
representation is composed of predefined operators and does not allow
one to define arbitrary, well known complex neighborhood operators.

The use of LS based solvers to solve CP problems occurs, for
instance, in SAT and MAX-SAT research, i.e.~the
WalkSat~\cite{selman_noise_1994} solver.  Van 
Hentenryck~\cite{van_hentenryck_constraint-based_2005} coined
the term Constraint-Based Local Search (CBLS) to describe LS systems
which derive the cost function from the CSP formulation of the problem.  
Typically, classical LS solvers use a generic neighborhood operator,
e.g. changing value of single variable guided by number of the
violated constraints.  Unfortunately, such strategies are in general
inferior to algorithms that exploit the specific problem
structure~\cite{hoos_local_2006}.  A simplified form of problem structure
knowledge was also explored in tools such as Adaptive
Search~\cite{truchet_2004}.  The other reason of their
inefficiency is a limited neighborhood size, which
may lead the solver to get stuck in a local optimum.  To overcome
those limitations, several metaheuristics have been proposed which
include some stochastic component or some form of
memory~\cite{hoos_local_2006}.

Large Neighborhood Search~\cite{pisinger_large_2010} (LNS) extends
popular exhaustive tree-search by interleaving the search with partial
relaxation of the problem.  This iterative process solves (using
tree-search) many relaxed versions of the problem, guiding the process
closer to the (local) optima.  Generally, in order to exploit the
problem's structure, one need to define the appropriate relaxation
method, but there exist several general strategies in wide use.  In
the most basic version, relaxation is a randomized process,
relaxing a random subset of variables.  In an adaptive
version, the solver changes the size of the relaxed subset
dynamically, balancing the computational cost with the size of the
induced neighborhood.  Propagation-Guided LNS~\cite{perron_propagation_2004} 
has
proved to be a competitive strategy, but even such generic strategies
have to be further parameterized to match a specific problem.  The
main advantage of LNS over previous methods is size of the
neighborhood, leading to a more diversified exploration.

During the past few years, CP/LS based solvers started gaining popularity.  In 
2016
the MiniZinc challenge~\cite{stuckey_minizinc_2014} for the first time
introduced the special category just for the LS solvers.  CBLS was
first implemented in the Comet system and currently is used in the
OscaR solver.  Due to the easy integration, LNS is widely adopted by
the most popular solvers, including Choco~\cite{prudhomme_choco_2017},
OscaR, or-tools~\cite{omme_or-tools_2014} and commercial ILOG CPLEX CP
Optimizer~\cite{noauthor_ibm_2014}.
MiniSearch~\cite{pesant_minisearch:_2015} allows to mix LNS with any
solver supporting the MiniZinc~\cite{bessiere_minizinc:_2007}
language. Yuck~\cite{marte_yuck_2017} and
OscaR/CBLS~\cite{bjordal_constraint-based_2015} provide dedicated
neighborhoods to handle specific global constraints.

\section{Goal of the research}
\label{sec:goal}
Recently, there has been noticeable progress in automated algorithm
configuration, including such complicated tasks as finding efficient
hyper-parameters~\cite{shahriari_taking_2016}, feature
selection~\cite{kaul_autolearn_2017}, selecting ASP solving strategy~\cite{hoos_claspfolio_2014, maratea_2014} and designing deep neural network
architectures~\cite{elsken_neural_2019}. Still there is no satisfying
method to create a problem-specific LS strategy. The presented research
aims at filling this gap by presenting a system capable of inferring
LS neighborhoods from a fine-grained problem representation
used in the Constraint Programming paradigm.

\begin{figure}[h]
    \centering
    \includegraphics[width=1.0\linewidth]{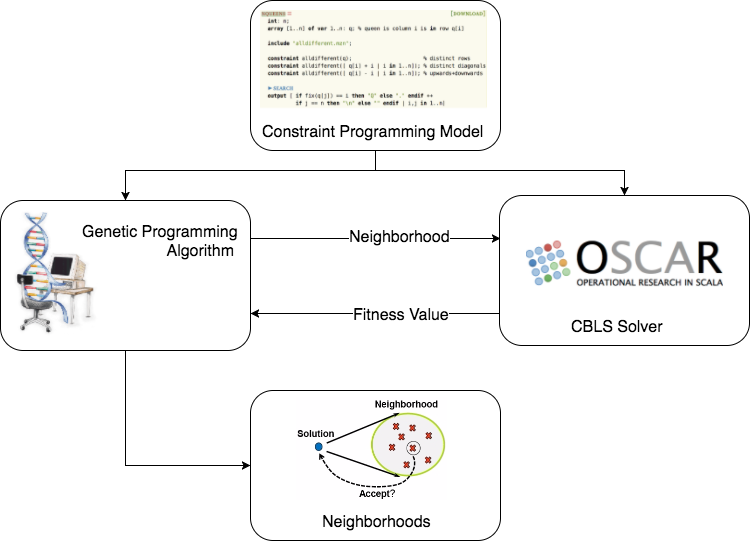}
    \caption[]{Genetic Programming approach to neighborhood synthesis with a 
    CP~Solver feedback loop.}
    \label{fig:gp-overview}
\end{figure}

The author argues that, given an appropriate representation of the neighborhood 
operators, it should be possible to automatically find specialized strategies by 
means of Genetic Programming techniques. Figure \ref{fig:gp-overview} presents a 
high-level view of such a system. The input model is used both the generate 
neighborhood candidates and to evaluate them with a Constraint Programming
solver. In order to build a similar system, the author has to answer at least three 
important questions:
\begin{itemize}
    \item How to calculate the fitness value of a neighborhood operator?
    \item What neighborhood representation should be used?
    \item What algorithm should be used to generate the neighborhood candidates?
\end{itemize}

\section{Current status of the research}

The hitherto conducted research has been focused on answering questions stated 
in Section \ref{sec:goal}. After studying the current state of the art, the author 
formulated a hypothesis that problem-specific neighborhood operators could be 
evaluated based on the impact over correctness of the solution. An efficient 
neighborhood operator would keep the correctness as an invariant, effectively 
pruning the search 
space from the inadmissible solutions. As the real-life problems may be too complex 
to find such neighborhood, the author has decided to use Constraint Programming 
model to decompose the considered problems based on the types 
of constraints used in their definition. Concluding, the neighborhood operator 
would be evaluated based on the number of constraint types it keeps satisfied.

Further works aimed at finding a proper neighborhood representation, capable 
of reasoning in terms of constraints and at the same time simple enough to be 
automatically processed. The research resulted in Neighborhood
Definition Language (NDL) --- a formal language capable of representing the
LS neighborhood operators in a fine-grained manner.
Compared to the general programming languages, NDL has a well defined and very 
simple semantics, limited to primitive recursion, effectively leading
to always terminating total programs. Compared to other declarative
approaches, it is much more self-contained and expressive enough to
represent even complex neighborhood operators, e.g. kempe chain or 3-opt.  Such 
results have been achieved partly with a rich problem representation borrowed from 
the Constraint Programming formulation, and partly with a limited set of
recursion schemes, effectively exploring problem's structure.

Recently, the author has experimentally verified his hypothesis with a synthesis 
framework based on the Grammar Evolution approach and a Prolog-based runtime, 
able to run and evaluate the NDL programs. Section \ref{sec:results} will present 
setup and results of the experiment in more detail.

\section{Preliminary results accomplished}
\label{sec:results}
At the moment of writing, two research papers have been accepted for presentation 
on the international conferences. First one~\cite{slazynski_towards_2019} is a short 
paper laying 
theoretical foundations for the future works. It formally presents the semantics and 
expressiveness of the NDL language, proving its capabilities with usage examples. 
The long version of this paper has been submitted for another conference and is 
being reviewed at the moment of writing.

\begin{figure}
    \centering
    \includegraphics[width=1\linewidth]{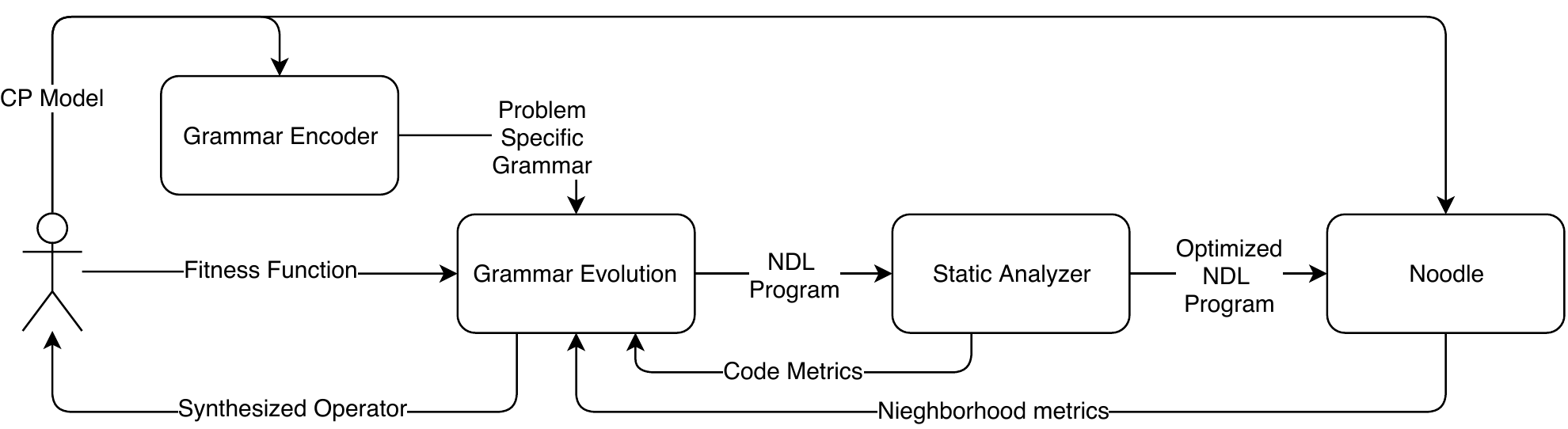}
    \caption{Architecture of the neighborhood synthesis framework.}
    \label{fig:experiment}
\end{figure}

Second publication~\cite{slazynski_generating_2019} is to be presented at the 
ICLP'2019 conference as a 
technical communication and covers the most recent experimental results. Figure 
\ref{fig:experiment} shows architecture of the implemented synthesis system called 
Noodle\footnote{See: \url{https://gitlab.geist.re/pro/ndl/}}, an 
instantiation of the high-level idea described in Section \ref{sec:goal}. Grammar 
Evolution, the synthesis algorithm used in the system, generates NDL programs 
based on a formal grammar, which is derived from the Constraint Programming 
model. The biggest advantage of this approach is that the synthesized programs 
are always syntactically correct and specifically refer to the structure of the 
considered problem. The synthesized programs are then evaluated in two stages. 
First a static analyzer checks quality of the program and optimizes the code. Then, 
all promising programs are applied to test data in order to check quality of the 
induced neighborhoods. Results of the evaluation are then fed back into the 
synthesizer, effectively closing the feedback loop. It is worth noting that NDL 
has been implemented as a domain specific language built on top of the SWI-Prolog 
system, mapping the NDL semantics directly to the SLDNF resolution. This choice 
was motivated by a built-in nondeterminism and meta-programming capabilities of 
the Prolog language. Results of the experiment are encouraging, the proposed 
system has been able to find complex neighborhood operators for the traveling 
salesman problem, three of them already known in the literature and two novel 
ones, requiring a further investigation into their effectiveness. Listing 
\ref{lst:2opt-opt} presents a 2-opt operator synthesized by the framework. 
Providing a step-by-step explanation of its workings would require introducing the 
NDL concepts in detail and goes beyond scope of this report. Since the first 
experiments, the framework has been also successfully applied to the graph 
coloring problem, reinventing the efficient and non-obvious kempe chain 
neighborhood.

\begin{listing}
    \begin{center}\hrule\end{center}
    \begin{alltt}
1.  constraint(all_diff_next, T0, T1) \(\land\)
2.  iterate(T3 - T4, T0, (
2.1.  constraint(all_diff_next, T4, T1) \(\land\)
2.2.  swap_values(T1, T0) \(\land\)
2.3.  swap_values(T4, T0)))
\end{alltt}
    \caption{2-opt Neighborhood operator synthesized by the Noodle framework.}
    \begin{center}\hrule\end{center}
    \label{lst:2opt-opt}
\end{listing}

\section{Open issues and expected achievements}

The biggest and still open challenge is integration of the proposed approach with 
the existing and commonly used tools. First of all, there has to be designed and 
implemented an algorithm to extract the problem structure from a Constraint 
Programming model defined in a general modeling language like MiniZinc. Currently, 
user has still to build the model manually with the Noodle DSL.

Furthermore, at the moment there is no proper LS solver integrated with 
the Noodle runtime. The author believes that it is crucial to create a new runtime, 
that 
could provide an existing solver with the synthesized neighborhood. A Constraint 
Guided Local Search solver, i.e. Oscar, seems to be the most promising target due 
to easily available Constraint Programming model. Another viable approach would 
consist in compiling NDL into OptaPlanner XML representation augmented with 
some external Java classes.

Solving the aforementioned issues would result in a user-friendly LS 
solver capable of inducing problem-specific neighborhoods. It would also enable to 
conduct a proper research on importance of such neighborhoods in various classes 
of problems, already modeled as MiniZinc models.

Other plans include experiments with the neighborhood representation, extending it 
with a new type of stochastic operators. The preliminary insight suggests that they 
could allow to define new interesting neighborhoods with no noticeable costs.

\bibliographystyle{eptcs}
\bibliography{bibliography/auto,bibliography/ndl,bibliography/problems,bibliography/misc,bibliography/cp_modelling,bibliography/search_modelling,bibliography/genetic_programming,bibliography/inductive_programming,bibliography/parallel_search}

\begin{thebibliography}{10}
\providecommand{\bibitemdeclare}[2]{}
\providecommand{\surnamestart}{}
\providecommand{\surnameend}{}
\providecommand{\urlprefix}{Available at }
\providecommand{\url}[1]{\texttt{#1}}
\providecommand{\href}[2]{\texttt{#2}}
\providecommand{\urlalt}[2]{\href{#1}{#2}}
\providecommand{\doi}[1]{doi:\urlalt{http://dx.doi.org/#1}{#1}}
\providecommand{\bibinfo}[2]{#2}

\bibitemdeclare{misc}{noauthor_optaplanner_nodate}
\bibitem{noauthor_optaplanner_nodate}
\emph{\bibinfo{title}{{OptaPlanner} {User} {Guide}}}.
\newblock
  \urlprefix\url{https://docs.optaplanner.org/7.15.0.Final/optaplanner-docs/html_single/index.html}.

\bibitemdeclare{misc}{noauthor_ibm_2014}
\bibitem{noauthor_ibm_2014}
 (\bibinfo{year}{2014}): \emph{\bibinfo{title}{{IBM} {CPLEX} {CP}
  {Optimizer}}}.
\newblock
  \urlprefix\url{http://www-01.ibm.com/software/commerce/optimization/cplex-cp-optimizer/}.

\bibitemdeclare{inproceedings}{akgun_framework_2018}
\bibitem{akgun_framework_2018}
\bibinfo{author}{{\"O}zg{\"u}r \surnamestart Akg{\"u}n\surnameend},
  \bibinfo{author}{Saad \surnamestart Attieh\surnameend},
  \bibinfo{author}{Ian~P. \surnamestart Gent\surnameend},
  \bibinfo{author}{Christopher \surnamestart Jefferson\surnameend},
  \bibinfo{author}{Ian \surnamestart Miguel\surnameend}, \bibinfo{author}{Peter
  \surnamestart Nightingale\surnameend}, \bibinfo{author}{Andr{\'a}s~Z.
  \surnamestart Salamon\surnameend}, \bibinfo{author}{Patrick \surnamestart
  Spracklen\surnameend} \& \bibinfo{author}{James \surnamestart
  Wetter\surnameend} (\bibinfo{year}{2018}): \emph{\bibinfo{title}{A Framework
  for Constraint Based Local Search using Essence}}.
\newblock In: {\sl \bibinfo{booktitle}{Proceedings of the Twenty-Seventh
  International Joint Conference on Artificial Intelligence, {IJCAI-18}}},
  \bibinfo{publisher}{International Joint Conferences on Artificial
  Intelligence Organization}, pp. \bibinfo{pages}{1242--1248},
  \doi{10.24963/ijcai.2018/173}.

\bibitemdeclare{inproceedings}{bjordal_declarative_2018}
\bibitem{bjordal_declarative_2018}
\bibinfo{author}{Gustav \surnamestart Bj{\"o}rdal\surnameend},
  \bibinfo{author}{Pierre \surnamestart Flener\surnameend},
  \bibinfo{author}{Justin \surnamestart Pearson\surnameend},
  \bibinfo{author}{Peter~J. \surnamestart Stuckey\surnameend} \&
  \bibinfo{author}{Guido \surnamestart Tack\surnameend} (\bibinfo{year}{2018}):
  \emph{\bibinfo{title}{Declarative {Local}-{Search} {Neighbourhoods} in
  {MiniZinc}}}.
\newblock In: {\sl \bibinfo{booktitle}{2018 {IEEE} 30th {International}
  {Conference} on {Tools} with {Artificial} {Intelligence} ({ICTAI})}}, pp.
  \bibinfo{pages}{98--105}, \doi{10.1109/ICTAI.2018.00025}.

\bibitemdeclare{article}{bjordal_constraint-based_2015}
\bibitem{bjordal_constraint-based_2015}
\bibinfo{author}{Gustav \surnamestart Bj{\"o}rdal\surnameend},
  \bibinfo{author}{Jean-No{\"e}l \surnamestart Monette\surnameend},
  \bibinfo{author}{Pierre \surnamestart Flener\surnameend} \&
  \bibinfo{author}{Justin \surnamestart Pearson\surnameend}
  (\bibinfo{year}{2015}): \emph{\bibinfo{title}{A constraint-based local search
  backend for {MiniZinc}}}.
\newblock {\sl \bibinfo{journal}{Constraints}}
  \bibinfo{volume}{20}(\bibinfo{number}{3}), pp. \bibinfo{pages}{325--345},
  \doi{10.1007/s10601-015-9184-z}.

\bibitemdeclare{article}{debeukelaer_james:_2017}
\bibitem{debeukelaer_james:_2017}
\bibinfo{author}{Herman \surnamestart De~Beukelaer\surnameend},
  \bibinfo{author}{Guy~F. \surnamestart Davenport\surnameend},
  \bibinfo{author}{Geert \surnamestart De~Meyer\surnameend} \&
  \bibinfo{author}{Veerle \surnamestart Fack\surnameend}
  (\bibinfo{year}{2017}): \emph{\bibinfo{title}{{JAMES}: {An} object-oriented
  {Java} framework for discrete optimization using local search
  metaheuristics}}.
\newblock {\sl \bibinfo{journal}{Software: Practice and Experience}}
  \bibinfo{volume}{47}(\bibinfo{number}{6}), pp. \bibinfo{pages}{921--938},
  \doi{10.1002/spe.2459}.

\bibitemdeclare{incollection}{elsken_neural_2019}
\bibitem{elsken_neural_2019}
\bibinfo{author}{Thomas \surnamestart Elsken\surnameend},
  \bibinfo{author}{Jan~Hendrik \surnamestart Metzen\surnameend} \&
  \bibinfo{author}{Frank \surnamestart Hutter\surnameend}
  (\bibinfo{year}{2019}): \emph{\bibinfo{title}{Neural Architecture Search}}.
\newblock In \bibinfo{editor}{Frank \surnamestart Hutter\surnameend},
  \bibinfo{editor}{Lars \surnamestart Kotthoff\surnameend} \&
  \bibinfo{editor}{Joaquin \surnamestart Vanschoren\surnameend}, editors: {\sl
  \bibinfo{booktitle}{Automated Machine Learning: Methods, Systems,
  Challenges}}, \bibinfo{publisher}{Springer International Publishing},
  \bibinfo{address}{Cham}, pp. \bibinfo{pages}{63--77},
  \doi{10.1007/978-3-030-05318-5\_3}.

\bibitemdeclare{article}{fourer_modeling_1990}
\bibitem{fourer_modeling_1990}
\bibinfo{author}{Robert \surnamestart Fourer\surnameend},
  \bibinfo{author}{David~M. \surnamestart Gay\surnameend} \&
  \bibinfo{author}{Brian~W. \surnamestart Kernighan\surnameend}
  (\bibinfo{year}{1990}): \emph{\bibinfo{title}{A {Modeling} {Language} for
  {Mathematical} {Programming}}}.
\newblock {\sl \bibinfo{journal}{Management Science}}
  \bibinfo{volume}{36}(\bibinfo{number}{5}), pp. \bibinfo{pages}{519--554},
  \doi{10.1287/mnsc.36.5.519}.
\newblock
  \urlprefix\url{https://pubsonline.informs.org/doi/abs/10.1287/mnsc.36.5.519}.

\bibitemdeclare{article}{frisch_essence:_2008}
\bibitem{frisch_essence:_2008}
\bibinfo{author}{Alan~M. \surnamestart Frisch\surnameend},
  \bibinfo{author}{Warwick \surnamestart Harvey\surnameend},
  \bibinfo{author}{Chris \surnamestart Jefferson\surnameend},
  \bibinfo{author}{Bernadette \surnamestart
  Mart{\'i}nez-Hern{\'a}ndez\surnameend} \& \bibinfo{author}{Ian \surnamestart
  Miguel\surnameend} (\bibinfo{year}{2008}): \emph{\bibinfo{title}{Essence: {A}
  constraint language for specifying combinatorial problems}}.
\newblock {\sl \bibinfo{journal}{Constraints}}
  \bibinfo{volume}{13}(\bibinfo{number}{3}), pp. \bibinfo{pages}{268--306},
  \doi{10.1007/s10601-008-9047-y}.

\bibitemdeclare{article}{gaspero_easylocal++:_2003}
\bibitem{gaspero_easylocal++:_2003}
\bibinfo{author}{Luca~Di \surnamestart Gaspero\surnameend} \&
  \bibinfo{author}{Andrea \surnamestart Schaerf\surnameend}
  (\bibinfo{year}{2003}): \emph{\bibinfo{title}{{EASYLOCAL}++: an
  object-oriented framework for the flexible design of local-search
  algorithms}}.
\newblock {\sl \bibinfo{journal}{Software: Practice and Experience}}
  \bibinfo{volume}{33}(\bibinfo{number}{8}), pp. \bibinfo{pages}{733--765},
  \doi{10.1002/spe.524}.
\newblock
  \urlprefix\url{https://onlinelibrary.wiley.com/doi/abs/10.1002/spe.524}.

\bibitemdeclare{book}{van_hentenryck_constraint-based_2005}
\bibitem{van_hentenryck_constraint-based_2005}
\bibinfo{author}{Pascal~Van \surnamestart Hentenryck\surnameend} \&
  \bibinfo{author}{Laurent \surnamestart Michel\surnameend}
  (\bibinfo{year}{2005}): \emph{\bibinfo{title}{Constraint-based local
  search}}.
\newblock \bibinfo{publisher}{{MIT} Press}.

\bibitemdeclare{article}{hoos_claspfolio_2014}
\bibitem{hoos_claspfolio_2014}
\bibinfo{author}{Holger~H. \surnamestart Hoos\surnameend},
  \bibinfo{author}{Marius~Thomas \surnamestart Lindauer\surnameend} \&
  \bibinfo{author}{Torsten \surnamestart Schaub\surnameend}
  (\bibinfo{year}{2014}): \emph{\bibinfo{title}{claspfolio 2: Advances in
  Algorithm Selection for Answer Set Programming}}.
\newblock {\sl \bibinfo{journal}{{TPLP}}}
  \bibinfo{volume}{14}(\bibinfo{number}{4-5}), pp. \bibinfo{pages}{569--585},
  \doi{10.1017/S1471068414000210}.

\bibitemdeclare{incollection}{hoos_local_2006}
\bibitem{hoos_local_2006}
\bibinfo{author}{Holger~H. \surnamestart Hoos\surnameend} \&
  \bibinfo{author}{Edward \surnamestart Tsang\surnameend}
  (\bibinfo{year}{2006}): \emph{\bibinfo{title}{Local {Search} {Methods}}}.
\newblock In: {\sl \bibinfo{booktitle}{Handbook of {Constraint}
  {Programming}}}, \bibinfo{series}{Foundations of {Artificial}
  {Intelligence}}, \bibinfo{publisher}{Elsevier Science Inc.},
  \bibinfo{address}{New York, NY, USA}, pp. \bibinfo{pages}{245--277},
  \doi{10.1016/S1574-6526(06)80009-X}.

\bibitemdeclare{inproceedings}{kaul_autolearn_2017}
\bibitem{kaul_autolearn_2017}
\bibinfo{author}{A.~\surnamestart Kaul\surnameend},
  \bibinfo{author}{S.~\surnamestart Maheshwary\surnameend} \&
  \bibinfo{author}{V.~\surnamestart Pudi\surnameend} (\bibinfo{year}{2017}):
  \emph{\bibinfo{title}{{AutoLearn} {\textemdash} {Automated} {Feature}
  {Generation} and {Selection}}}.
\newblock In: {\sl \bibinfo{booktitle}{2017 {IEEE} {International} {Conference}
  on {Data} {Mining} ({ICDM})}}, pp. \bibinfo{pages}{217--226},
  \doi{10.1109/ICDM.2017.31}.

\bibitemdeclare{incollection}{landtsheer_combining_2018}
\bibitem{landtsheer_combining_2018}
\bibinfo{author}{Renaud~De \surnamestart Landtsheer\surnameend},
  \bibinfo{author}{Yoann \surnamestart Guyot\surnameend},
  \bibinfo{author}{Gustavo \surnamestart Ospina\surnameend} \&
  \bibinfo{author}{Christophe \surnamestart Ponsard\surnameend}
  (\bibinfo{year}{2018}): \emph{\bibinfo{title}{Combining {Neighborhoods} into
  {Local} {Search} {Strategies}}}.
\newblock In: {\sl \bibinfo{booktitle}{Recent {Developments} in
  {Metaheuristics}}}, \bibinfo{series}{Operations {Research}/{Computer}
  {Science} {Interfaces} {Series}}, \bibinfo{publisher}{Springer, Cham}, pp.
  \bibinfo{pages}{43--57}, \doi{10.1007/978-3-319-58253-5\_3}.

\bibitemdeclare{article}{maratea_2014}
\bibitem{maratea_2014}
\bibinfo{author}{Marco \surnamestart Maratea\surnameend}, \bibinfo{author}{Luca
  \surnamestart Pulina\surnameend} \& \bibinfo{author}{Francesco \surnamestart
  Ricca\surnameend} (\bibinfo{year}{2014}): \emph{\bibinfo{title}{A
  multi-engine approach to answer-set programming}}.
\newblock {\sl \bibinfo{journal}{{TPLP}}}
  \bibinfo{volume}{14}(\bibinfo{number}{6}), pp. \bibinfo{pages}{841--868},
  \doi{10.1017/S1471068413000094}.

\bibitemdeclare{misc}{marte_yuck_2017}
\bibitem{marte_yuck_2017}
\bibinfo{author}{Michael \surnamestart Marte\surnameend}
  (\bibinfo{year}{2017}): \emph{\bibinfo{title}{Yuck is a constraint-based
  local-search solver with {FlatZinc} interface}}.
\newblock \urlprefix\url{https://github.com/informarte/yuck}.
\newblock \bibinfo{note}{Accessed 2019/05/13}.

\bibitemdeclare{inproceedings}{michel_localizer_1997}
\bibitem{michel_localizer_1997}
\bibinfo{author}{Laurent \surnamestart Michel\surnameend} \&
  \bibinfo{author}{Pascal \surnamestart Van~Hentenryck\surnameend}
  (\bibinfo{year}{1997}): \emph{\bibinfo{title}{Localizer {A} modeling language
  for local search}}.
\newblock In \bibinfo{editor}{Gert \surnamestart Smolka\surnameend}, editor:
  {\sl \bibinfo{booktitle}{Principles and {Practice} of {Constraint}
  {Programming}-{CP}97}}, \bibinfo{series}{Lecture {Notes} in {Computer}
  {Science}}, \bibinfo{publisher}{Springer Berlin Heidelberg}, pp.
  \bibinfo{pages}{237--251}, \doi{10.1007/BFb0017443}.

\bibitemdeclare{incollection}{bessiere_minizinc:_2007}
\bibitem{bessiere_minizinc:_2007}
\bibinfo{author}{Nicholas \surnamestart Nethercote\surnameend},
  \bibinfo{author}{Peter~J. \surnamestart Stuckey\surnameend},
  \bibinfo{author}{Ralph \surnamestart Becket\surnameend},
  \bibinfo{author}{Sebastian \surnamestart Brand\surnameend},
  \bibinfo{author}{Gregory~J. \surnamestart Duck\surnameend} \&
  \bibinfo{author}{Guido \surnamestart Tack\surnameend} (\bibinfo{year}{2007}):
  \emph{\bibinfo{title}{{MiniZinc}: {Towards} a {Standard} {CP} {Modelling}
  {Language}}}.
\newblock In \bibinfo{editor}{Christian \surnamestart Bessi{\`e}re\surnameend},
  editor: {\sl \bibinfo{booktitle}{Principles and {Practice} of {Constraint}
  {Programming} {\textendash} {CP} 2007}}, \bibinfo{volume}{4741},
  \bibinfo{publisher}{Springer Berlin Heidelberg}, \bibinfo{address}{Berlin,
  Heidelberg}, pp. \bibinfo{pages}{529--543},
  \doi{10.1007/978-3-540-74970-7\_38}.

\bibitemdeclare{techreport}{omme_or-tools_2014}
\bibitem{omme_or-tools_2014}
\bibinfo{author}{Nikolaj~van \surnamestart Omme\surnameend},
  \bibinfo{author}{Laurent \surnamestart Perron\surnameend} \&
  \bibinfo{author}{Vincent \surnamestart Furnon\surnameend}
  (\bibinfo{year}{2014}): \emph{\bibinfo{title}{or-tools user's manual}}.
\newblock \bibinfo{type}{Technical Report}, \bibinfo{institution}{Google}.

\bibitemdeclare{book}{oscar_team_oscar:_2012}
\bibitem{oscar_team_oscar:_2012}
\bibinfo{author}{\surnamestart {OscaR Team}\surnameend} (\bibinfo{year}{2012}):
  \emph{\bibinfo{title}{{OscaR}: {Scala} in {OR}}}.

\bibitemdeclare{inproceedings}{perron_propagation_2004}
\bibitem{perron_propagation_2004}
\bibinfo{author}{Laurent \surnamestart Perron\surnameend},
  \bibinfo{author}{Paul \surnamestart Shaw\surnameend} \&
  \bibinfo{author}{Vincent \surnamestart Furnon\surnameend}
  (\bibinfo{year}{2004}): \emph{\bibinfo{title}{Propagation {Guided} {Large}
  {Neighborhood} {Search}}}.
\newblock In: {\sl \bibinfo{booktitle}{Principles and {Practice} of
  {Constraint} {Programming} {\textendash} {CP} 2004}},
  \bibinfo{series}{Lecture {Notes} in {Computer} {Science}},
  \bibinfo{publisher}{Springer, Berlin, Heidelberg}, pp.
  \bibinfo{pages}{468--481}, \doi{10.1007/978-3-540-30201-8\_35}.

\bibitemdeclare{incollection}{pisinger_large_2010}
\bibitem{pisinger_large_2010}
\bibinfo{author}{David \surnamestart Pisinger\surnameend} \&
  \bibinfo{author}{Stefan \surnamestart R{\o}pke\surnameend}
  (\bibinfo{year}{2010}): \emph{\bibinfo{title}{Large {Neighborhood}
  {Search}}}.
\newblock In: {\sl \bibinfo{booktitle}{Handbook of {Metaheuristics}}},
  \bibinfo{publisher}{Springer}, pp. \bibinfo{pages}{399--420},
  \doi{10.1016/j.ejor.2004.08.015}.

\bibitemdeclare{book}{prudhomme_choco_2017}
\bibitem{prudhomme_choco_2017}
\bibinfo{author}{Charles \surnamestart Prud'homme\surnameend},
  \bibinfo{author}{Jean-Guillaume \surnamestart Fages\surnameend} \&
  \bibinfo{author}{Xavier \surnamestart Lorca\surnameend}
  (\bibinfo{year}{2017}): \emph{\bibinfo{title}{Choco {Documentation}}}.
\newblock \bibinfo{publisher}{TASC - LS2N CNRS UMR 6241, COSLING S.A.S.}
\newblock \urlprefix\url{http://www.choco-solver.org}.

\bibitemdeclare{incollection}{pesant_minisearch:_2015}
\bibitem{pesant_minisearch:_2015}
\bibinfo{author}{Andrea \surnamestart Rendl\surnameend}, \bibinfo{author}{Tias
  \surnamestart Guns\surnameend}, \bibinfo{author}{Peter~J. \surnamestart
  Stuckey\surnameend} \& \bibinfo{author}{Guido \surnamestart Tack\surnameend}
  (\bibinfo{year}{2015}): \emph{\bibinfo{title}{{MiniSearch}: {A}
  {Solver}-{Independent} {Meta}-{Search} {Language} for {MiniZinc}}}.
\newblock In \bibinfo{editor}{Gilles \surnamestart Pesant\surnameend}, editor:
  {\sl \bibinfo{booktitle}{Principles and {Practice} of {Constraint}
  {Programming}}}, \bibinfo{volume}{9255}, \bibinfo{publisher}{Springer
  International Publishing}, \bibinfo{address}{Cham}, pp.
  \bibinfo{pages}{376--392}, \doi{10.1007/978-3-319-23219-5\_27}.

\bibitemdeclare{inproceedings}{schaerf_local++:_1999}
\bibitem{schaerf_local++:_1999}
\bibinfo{author}{Andrea \surnamestart Schaerf\surnameend},
  \bibinfo{author}{Maurizio \surnamestart Lenzerini\surnameend} \&
  \bibinfo{author}{Marco \surnamestart Cadoli\surnameend}
  (\bibinfo{year}{1999}): \emph{\bibinfo{title}{{LOCAL}++: a {C}++ framework
  for local search algorithms}}.
\newblock In: {\sl \bibinfo{booktitle}{Proceedings {Technology} of
  {Object}-{Oriented} {Languages} and {Systems}. {TOOLS} 29 ({Cat}.
  {No}.{PR}00275)}}, pp. \bibinfo{pages}{152--161},
  \doi{10.1109/TOOLS.1999.779008}.

\bibitemdeclare{inproceedings}{selman_noise_1994}
\bibitem{selman_noise_1994}
\bibinfo{author}{Bart \surnamestart Selman\surnameend},
  \bibinfo{author}{Henry~A. \surnamestart Kautz\surnameend} \&
  \bibinfo{author}{Bram \surnamestart Cohen\surnameend} (\bibinfo{year}{1994}):
  \emph{\bibinfo{title}{Noise Strategies for Improving Local Search}}.
\newblock In \bibinfo{editor}{Barbara \surnamestart Hayes{-}Roth\surnameend} \&
  \bibinfo{editor}{Richard~E. \surnamestart Korf\surnameend}, editors: {\sl
  \bibinfo{booktitle}{Proceedings of the 12th National Conference on Artificial
  Intelligence, Seattle, WA, USA, July 31 - August 4, 1994, Volume 1.}},
  \bibinfo{publisher}{{AAAI} Press / The {MIT} Press}, pp.
  \bibinfo{pages}{337--343}.
\newblock \urlprefix\url{http://www.aaai.org/Library/AAAI/1994/aaai94-051.php}.

\bibitemdeclare{article}{shahriari_taking_2016}
\bibitem{shahriari_taking_2016}
\bibinfo{author}{B.~\surnamestart Shahriari\surnameend},
  \bibinfo{author}{K.~\surnamestart Swersky\surnameend},
  \bibinfo{author}{Z.~\surnamestart Wang\surnameend}, \bibinfo{author}{R.~P.
  \surnamestart Adams\surnameend} \& \bibinfo{author}{N.~de \surnamestart
  Freitas\surnameend} (\bibinfo{year}{2016}): \emph{\bibinfo{title}{Taking the
  {Human} {Out} of the {Loop}: {A} {Review} of {Bayesian} {Optimization}}}.
\newblock {\sl \bibinfo{journal}{Proceedings of the IEEE}}
  \bibinfo{volume}{104}(\bibinfo{number}{1}), pp. \bibinfo{pages}{148--175},
  \doi{10.1109/JPROC.2015.2494218}.

\bibitemdeclare{inproceedings}{slazynski_generating_2019}
\bibitem{slazynski_generating_2019}
\bibinfo{author}{Mateusz \surnamestart \'{S}la\.{z}y\'{n}ski\surnameend},
  \bibinfo{author}{Salvador \surnamestart Abreu\surnameend} \&
  \bibinfo{author}{Grzegorz~J. \surnamestart Nalepa\surnameend}
  (\bibinfo{year}{2019}): \emph{\bibinfo{title}{Generating {Local} {Search}
  {Neighborhood} with {Synthesized} {Logic} {Programs}}}.
\newblock In: {\sl \bibinfo{booktitle}{{{ICLP} 2019}, Electronic {Proceedings}
  in {Theoretical} {Computer} {Science}}}, \bibinfo{address}{Las Cruces, New
  Mexico, USA}.
\newblock \bibinfo{note}{Forthcoming}.

\bibitemdeclare{inproceedings}{slazynski_towards_2019}
\bibitem{slazynski_towards_2019}
\bibinfo{author}{Mateusz \surnamestart \'{S}la\.{z}y\'{n}ski\surnameend},
  \bibinfo{author}{Salvador \surnamestart Abreu\surnameend} \&
  \bibinfo{author}{Grzegorz~J. \surnamestart Nalepa\surnameend}
  (\bibinfo{year}{2019}): \emph{\bibinfo{title}{Towards a Formal Specification
  of Local Search Neighborhoods from a Constraint Satisfaction Problem
  Structure}}.
\newblock In: {\sl \bibinfo{booktitle}{Proceedings of the Genetic and
  Evolutionary Computation Conference Companion}}, \bibinfo{series}{GECCO '19},
  \bibinfo{publisher}{ACM}, \bibinfo{address}{New York, NY, USA}, pp.
  \bibinfo{pages}{137--138}, \doi{10.1145/3319619.3321968}.

\bibitemdeclare{article}{stuckey_minizinc_2014}
\bibitem{stuckey_minizinc_2014}
\bibinfo{author}{Peter~J. \surnamestart Stuckey\surnameend},
  \bibinfo{author}{Thibaut \surnamestart Feydy\surnameend},
  \bibinfo{author}{Andreas \surnamestart Schutt\surnameend},
  \bibinfo{author}{Guido \surnamestart Tack\surnameend} \&
  \bibinfo{author}{Julien \surnamestart Fischer\surnameend}
  (\bibinfo{year}{2014}): \emph{\bibinfo{title}{The {MiniZinc} {Challenge}
  2008{\textendash}2013}}.
\newblock {\sl \bibinfo{journal}{AI Magazine}}
  \bibinfo{volume}{35}(\bibinfo{number}{2}), pp. \bibinfo{pages}{55--60},
  \doi{10.1609/aimag.v35i2.2539}.

\bibitemdeclare{article}{truchet_2004}
\bibitem{truchet_2004}
\bibinfo{author}{Charlotte \surnamestart Truchet\surnameend} \&
  \bibinfo{author}{Philippe \surnamestart Codognet\surnameend}
  (\bibinfo{year}{2004}): \emph{\bibinfo{title}{Musical constraint satisfaction
  problems solved with adaptive search}}.
\newblock {\sl \bibinfo{journal}{Soft Comput.}}
  \bibinfo{volume}{8}(\bibinfo{number}{9}), pp. \bibinfo{pages}{633--640},
  \doi{10.1007/s00500-004-0389-0}.

\bibitemdeclare{book}{van_hentenryck_opl_1999}
\bibitem{van_hentenryck_opl_1999}
\bibinfo{author}{Pascal \surnamestart Van~Hentenryck\surnameend}
  (\bibinfo{year}{1999}): \emph{\bibinfo{title}{The {OPL} {Optimization}
  {Programming} {Language}}}.
\newblock \bibinfo{publisher}{MIT Press}, \bibinfo{address}{Cambridge, MA,
  USA}.

\end{thebibliography}
\end{document}